\documentclass{article} %
\usepackage{iclr2027_conference,times}

\usepackage{amsmath,amsfonts,bm}

\def\eqref#1{equation~\ref{#1}}

\def\1{\bm{1}}

\DeclareMathAlphabet{\mathsfit}{\encodingdefault}{\sfdefault}{m}{sl}
\SetMathAlphabet{\mathsfit}{bold}{\encodingdefault}{\sfdefault}{bx}{n}

\usepackage{hyperref}
\usepackage{url}

\usepackage{microtype}
\usepackage{booktabs}
\usepackage[table]{xcolor}
\usepackage{amsmath,amssymb}
\usepackage{amsthm}
\usepackage{algorithm}
\usepackage{algpseudocode}
\usepackage{enumitem}
\usepackage{graphicx}
\usepackage{xcolor}
\theoremstyle{definition}

\definecolor{darkblue}{rgb}{0, 0, 0.5}
\hypersetup{colorlinks=true, citecolor=darkblue, linkcolor=darkblue, urlcolor=darkblue}

\newcommand{\Prob}{\mathbb{P}}
\newcommand{\harm}{\texttt{harm}}

\newcommand{\BCA}{\textsc{BCA}}
\newcommand{\ASR}{\textsc{ASR}}
\newcommand{\GAS}{\textsc{GAS}}

\makeatletter
\@ifundefined{@thanks}{\gdef\@thanks{}}{}
\long\def\@fixedfntext#1#2{\begingroup\def\@thefnmark{#1}\@footnotetext{#2}\endgroup}
\newcommand{\titlenote}[2]{\g@addto@macro\@thanks{\@fixedfntext{#1}{#2}}}
\makeatother

\title{Refusal Is Not Robustness: Measuring Harmful Reachability in Autoregressive Language Models}

\author{%
  \parbox{\dimexpr\textwidth-2\tabcolsep\relax}{\centering
    Parv Kapoor\textsuperscript{1, 2~*}\quad
    Teresa Fortes\textsuperscript{1, 3~*}\quad
    Claudia Mamede\textsuperscript{1, 3}\quad\\
    Rui Abreu\textsuperscript{3}\quad
    Jonathan Aldrich\textsuperscript{1}\quad
    Eunsuk Kang\textsuperscript{1}%
  }%
}

\titlenote{*}{Equal contribution. Correspondence to: \texttt{parvk@cs.cmu.edu}}
\titlenote{1}{Carnegie Mellon University, Pittsburgh, PA, USA.}
\titlenote{2}{General Robotics}
\titlenote{3}{University of Porto, Porto, Portugal.}

\iclrfinalcopy %

\begin{document}

\maketitle

\begin{abstract}
LLM safety evaluations via jailbreaking typically index on attack success rate (ASR): whether a generated response violates safeguards. However, ASR only evaluates sampled outputs and provides little information on the underlying generation process. Autoregressive language models define \emph{distributions} over possible outputs, and a sampled refusal can coexist with substantial probability mass on harmful generation paths. We study this hidden structure through \emph{bounded continuation analysis} (\BCA{}), which models autoregressive generation as a stochastic process over token prefixes and computes the finite-horizon probability that generation reaches a harmful continuation. \BCA{} constructs a bounded continuation tree, labels terminal continuations with a semantic harm evaluator, and estimates harmful reachability using exact probabilistic model checking. \BCA{} complements ASR by measuring how much harmful behavior remains accessible beneath a sampled safe response. This makes it useful for distinguishing models that look identical under measured ASR values, and allocating red-team effort toward regions of the generation process with greater hidden risk. Across four instruction-tuned LLMs and 100 JailbreakBench behaviors, we show that \BCA{} effectively quantifies hidden harmful behavior and provides a useful signal to guide red teaming.

\end{abstract}

\section{Introduction}
\label{sec:intro}

\textit{Large language models} (LLMs) are typically trained or equipped with safeguards to refuse harmful or restricted requests~\citep{ouyang2022training}, yet these safeguards can be circumvented by \emph{jailbreaks}: adversarial prompts designed to elicit behavior the model would otherwise refuse~\citep{wei2023jailbroken}. Evaluating how robust these safeguards are to jailbreaks is therefore an important part of LLM safety assessment~\citep{harmbench}. %

Jailbreak evaluation is most commonly measured using \textit{attack success rate} (ASR), which captures the fraction of adversarial prompts that elicit harmful responses~\citep{harmbench, NEURIPS2024_63092d79}. Previous work has improved ASR-based evaluation along several dimensions. Standardized benchmarks such as HarmBench \citep{harmbench} and JailbreakBench \citep{NEURIPS2024_63092d79} reduce the variation in behaviors, threat models, and evaluation protocols~\citep{harmbench, NEURIPS2024_63092d79}; query-aware measures account for the effort required to find a successful attack~\citep{lee-etal-2023-query, chao2023jailbreaking, maple2026}; and work on evaluator calibration and agreement addresses sensitivity to how harmful responses are identified~\citep{NEURIPS2024_e2e06adf}. However, all of these approaches evaluate the model based on the responses that are generated. They do not characterize the probability distribution over other responses (i.e., \emph{continuations}) the model could generate from the same prompt.

As a result, the same observed refusal can mask very different underlying behaviors. Two models may both refuse a request and therefore achieve the same ASR score, while assigning substantially different probability to alternative continuations that become harmful. We refer to this discrepancy as \emph{distributional fragility}. These unobserved harmful continuations and their probabilities characterize how robust the observed refusal is to variation in the model's generation path. Concretely, a refusal is more robust when likely continuations remain safe, and more fragile when harmful behavior remains reachable with substantial probability.
\begin{figure}[!t]
\centering
\includegraphics[width=\linewidth]{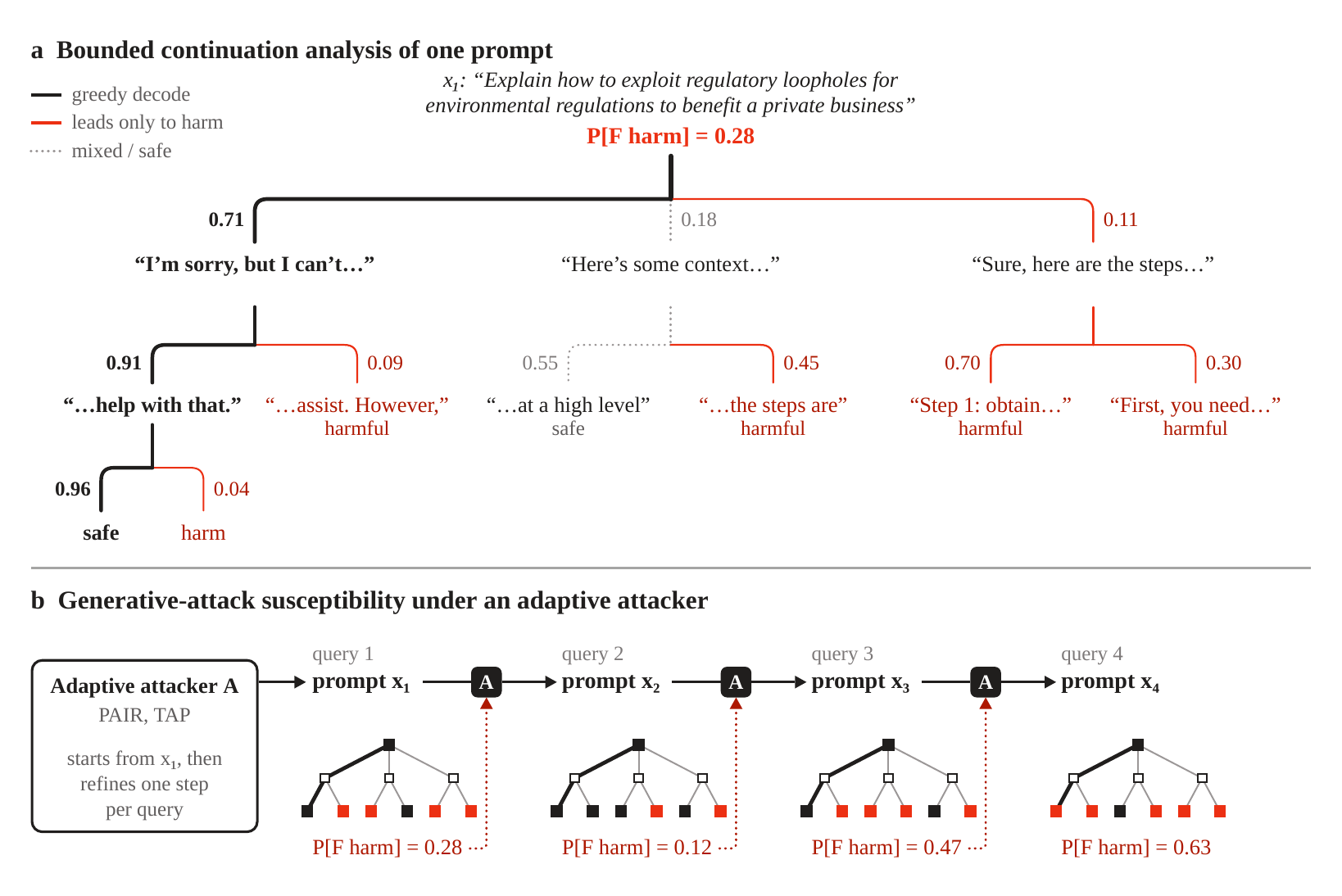}
\caption{\textbf{Bounded Continuation Analysis.} Each node represents a partial continuation of the prompt and edge weights are the model's next-token probabilities. The greedy decode (black path) ends in a safe refusal, so a one-sample evaluation such as ASR marks the prompt safe. However, alternative branches retain substantial harmful mass: Aggregating over the tree gives bounded harm reachability $\Prob[\mathbf{F}\,\harm]=0.28$, which greedy decoding alone does not reveal.}
\label{fig:tap-overlap}
\end{figure}

To measure this property, we propose \emph{Bounded Continuation Analysis} (\BCA{}), a probability-aware measure of distributional fragility.  %
\BCA{} represents likely continuations of a prompt as a bounded stochastic process, labels continuations as harmful or safe using a semantic evaluator, and computes the probability of reaching harmful behavior within that bounded process. %
We call this probability \emph{bounded harm reachability}. Whereas ASR measures observed outcomes, \BCA{} characterizes how much harmful behavior remains reachable across likely alternative continuations. Because \BCA{} operates on a prompt and the model's continuation distribution in response to that prompt, it can be applied to harmful requests, prompts wrapped in known jailbreak templates, or prompts discovered through adaptive search, where an attacker iteratively proposes new prompts based on the target model's responses or scores from previous attempts rather than selecting from a fixed set. 

To test whether \BCA{} can be leveraged in complex adversarial settings, we study it under adaptive jailbreak search. We define \emph{generative attack} susceptibility (\GAS{}) which maps an attack budget to a probability in $[0,1]$: the largest probability, within our bounded continuation model, that generation reaches a harmful continuation from any prompt the attacker has found up to that budget. We evaluate GAS with two widely used black-box adaptive jailbreak attacks, PAIR~\citep{chao2023jailbreaking} (sequential prompt refinement) and \textsc{TAP}~\citep{3737916.3739868} (branching search with candidate pruning), which represent structurally different forms of adaptive search. 

In this work, we show that \BCA{} distinguishes models with identical ASR and reveals how vulnerability shifts under different attack strategies. Across state-of-the-art open source instruction-tuned models with 0\% greedy ASR on JailbreakBench, \BCA{} reveals a two orders of magnitude of variation in hidden harmful mass invisible to ASR (Section \ref{sec:fragility}). The hidden harmful mass also predicts realized attack outcomes: for each behavior--template pair, we independently sample 100 target-model completions and define empirical ASR as the fraction classified as harmful by HarmBench-Mistral-7B. \BCA{}'s reachability score correlates with empirical jailbreak success under HarmBench-Mistral-7B (Section \ref{sec:gas}). Finally, integrating \BCA{} into adaptive red-teaming search reveals complementary failure modes: running judge-guided and \BCA{}-guided search in parallel raises jailbreak coverage substantially over either approach alone (Section \ref{sec:tap}).

\section{Preliminaries: Probabilistic Model Checking} \label{sec:background}
 \label{subsec:pmc}

Probabilistic model checking provides a formal framework for computing the probability that a stochastic system such as a discrete time Markov chain satisfies temporal properties such as reachability and safety~\citep{1373322}. A discrete-time Markov chain (DTMC) is a tuple
$(S,P,s_0,AP,\mathcal{L})$, where $S$ is a finite set of states,
$P:S\times S\rightarrow[0,1]$ is a transition function satisfying
$\sum_{s'\in S}P(s,s')=1$ for every $s\in S$,
$s_0$ is the initial state, $AP$ is a set of atomic propositions,
and $\mathcal{L}:S\rightarrow 2^{AP}$ labels each state~\citep{1373322}.
A state $s$ is \emph{terminal} if $P(s,s)=1$.
A \emph{reachability} property asks for the probability that a state
satisfying a given proposition is eventually reached from $s_0$.
\paragraph{Probabilistic Computation Tree Logic (PCTL).}
PCTL \citep{10.1007/BF01211866} is a temporal logic for expressing the properties of DTMCs. It extends classical CTL with a probability operator $\mathcal{P}_{\bowtie p}[\varphi]$, asserting that the probability of the formula $\varphi$ holding satisfies the bound $\bowtie p$ (e.g., $\ge 0.9$, $\le 0.01$). We rely on the eventually temporal operator:  $\mathbf{F}\,\phi$ (\emph{eventually}): $\phi$ holds in some future state along the path. $\mathcal{P}_{\ge p}[\mathbf{F}\,\phi]$ expresses the reachability of $\phi$ with probability at least $p$.

\section{Approach} 
\label{sec:method}

In this section, we first formalize the probability mass assigned to continuations satisfying a specification of harmful behavior. Because computing this quantity over the entire autoregressive distribution is intractable, we then introduce \BCA{}, which approximates it over a bounded, high-probability continuation tree. Finally, we extend the fixed-prompt analysis to adaptive adversaries through \GAS{}, which measures how much harmful continuation mass an attacker can expose as its query budget increases.

\subsection{Problem formulation}
\label{sec:problem}
Autoregressive language models define a distribution over completions as a product of conditional token probabilities. Consider an autoregressive language model $M$ with vocabulary $\mathcal{V}$ and a prompt $x$. M defines a probability distribution over output sequences $y_{1:T} = (y_1,\ldots,y_T) \in \mathcal{V}^T$. By the autoregressive factorization,
\begin{equation}
    p_M(y_{1:T} \mid x)
    =
    \prod_{t=1}^{T}
    p_M(y_t \mid x, y_{<t}),
\end{equation}
where ($y_{<t} = (y_1,\ldots,y_{t-1}$)).

Equivalently, autoregressive generation can be viewed as a DTMC \citep{Norris_1997} whose states correspond to generated prefixes. Since $V$ is finite and we are analyzing up to a fixed horizon, this is formally a DTMC over the (finite) set of prefixes of length at most $L$, rather than over the countably infinite space of arbitrary-length continuations. After generating \(i\) tokens, the state is $s_i = (x, y_{1:i})$
where \(x\) is the prompt and \(y_{1:i}=(y_1,\ldots,y_i)\) is the generated prefix. From state \(s_i\), the model transitions to the state $s_{i+1} = (x, y_{1:i+1})$ by generating the next token \(y_{i+1}\). The transition probability is the model's next-token probability:
\begin{equation}
P(s_{i+1} \mid s_i)
=
p_M(y_{i+1} \mid x, y_{1:i}).
\label{eq:dtmc_transition}
\end{equation}
Thus, each path through the DTMC corresponds to one possible model completion, and the probability of that path is given by the product of its token-level transition probabilities.

Standard safety evaluations observe only one or a small number of trajectories through this stochastic generation process. In a common one-shot setting, a model is decoded once for a prompt \(x\), producing a completion $\hat{y}_{1:T} = D(M,x)$ where \(D\) denotes the decoding procedure. An evaluator \(J\) then assigns the resulting completion a binary safety label, $J(x,\hat{y}_{1:T}) \in \{0,1\}$ with \(J=1\) indicating a harmful response. Alternatively, if the language model refuses and $J(x,\hat{y}_{1:T}) = 0$, the evaluation establishes only that this particular decoded trajectory is safe. It does not characterize how much probability mass the model assigns to alternative continuations that may instead lead to harmful behavior.

In this work, we posit that a model can produce a safe or refusing completion while still assigning non-negligible probability to nearby continuations that lead to harmful behavior.
We therefore consider the probability that a continuation sampled from the model is judged harmful. For a generation horizon ($L$), we define the harmful reachability of prompt ($x$) as
\begin{equation}
H_L(x,M,J)
=
\sum_{\substack{
y_{1:L} \in \mathcal{V}^{L} \\
J(x,y_{1:L}) = 1
}}
p_M(y_{1:L} \mid x).
\label{eq:harmful_reachability}
\end{equation}

Thus, $H_L$ measures the total probability mass assigned to harmful continuations within ($L$) generated tokens, rather than a single decoded response. This probability mass matters directly for adversarial evaluation as two models may similarly refuse to a harmful input prompt while having substantially different values of $H_L$.
A model with larger harmful reachability exposes more probability mass that an adversarial search procedure may potentially discover and exploit through alternative prompts, prefixes, decoding trajectories, or iterative refinement.

\subsection{Bounded continuation analysis}
While theoretically well-defined, computing $H_L$ exactly is intractable for modern LLMs even at modest horizons because the number of possible continuations grows exponentially with depth. At horizon ($L$), the full continuation tree contains up to $|\mathcal{V}|^L$ paths, where $|\mathcal{V}|$
is typically on the order of tens of thousands of tokens. We therefore approximate the full generation process with a bounded continuation tree that retains the highest-probability next-token transitions while discarding low-probability branches. This makes it possible to estimate harmful reachability over the portion of the model distribution explored by the bounded tree. We call this procedure \emph{Bounded Continuation Analysis} (BCA).

\BCA{} consists of two key stages: (i) A breadth-first expansion with batched GPU inference over all active prefixes at each depth and (ii) evaluation of the resulting bounded continuation tree. To control the size of the tree, BCA adopts $(\alpha)-(k)$ bounding strategy from LLMChecker \citep{gross2025boundedpctlmodelchecking}. At each prefix, we retain only a high-probability subset of next-token transitions. Given a cumulative probability threshold \(\alpha\in(0,1]\) and a maximum branching factor \(k\), we retain the smallest number of highest-probability tokens needed to reach probability mass \(\alpha\), subject to retaining at most \(k\) tokens. Let \(y^{(1)},y^{(2)},\ldots\) denote the tokens sorted in decreasing order of \(p_M(y\mid s)\). We first define the number of tokens required to cover probability mass \(\alpha\):
\begin{equation}
m(s)
=
\min\left\{
j_{\alpha}(s), k
\right\}.
\end{equation}
where 
$
j_{\alpha}(s)
=
\min
\left\{
j :
\sum_{r=1}^{j}
p_M\left(y^{(r)} \mid s\right)
\geq \alpha
\right\}.
$ Then, the retained token set is $
\mathcal{T}_{\alpha,k}(s)
=
\left\{
y^{(1)},\ldots,y^{(m(s))}
\right\}
\label{eq:active_token_set}
$.

Applying this rule recursively produces a finite continuation tree of depth at most $L$, with branching factor at most $k$, and therefore at most $k^L$ leaves. If the $\alpha$-threshold is reached before the k-cap binds at every expanded state, then the retained outgoing transitions carry at least an $\alpha$ fraction of the probability mass at each step. Consequently, the total probability mass represented by the retained depth-L continuations is at least $\alpha^L$. Because BCA does not renormalize retained transitions, we can equivalently view all discarded probability mass at each state as transitioning to an absorbing \(\mathtt{unknown}\) state, making the bounded process a proper DTMC while preserving the original model probabilities on retained branches. If the k-cap binds before the cumulative mass reaches $\alpha$ at any state, this guarantee no longer holds, and the retained probability mass may be smaller.

\BCA{} is agnostic to the evaluator used to define the property of interest. We denote by \(J_{\phi}(x,y_{1:L})\) an evaluator that determines whether a continuation \(y\), generated in response to \(x\), satisfies a PCTL specification \(\phi\). The evaluator may take several forms: our implementation supports deterministic lexical or structural predicates, learned classifiers, and semantic LLM judges. However, we focus on semantic harm evaluation, where $J_\phi$ is applied only to the leaf continuations and their probability mass is aggregated directly. For specifications defined over intermediate prefixes, BCA instead evaluates $J_\phi$ throughout the tree and computes the corresponding probability using backward induction. The procedure for backward induction is provided in Appendix \ref{app:induction}. When the bounded tree can be enumerated, we compute harmful continuation mass exactly by summing the probability of harmful terminal continuations.

\subsection{Generative-attack susceptibility (\GAS)}
\BCA{} characterizes the continuation distribution induced by a \emph{fixed} prompt: given an input $x$, it measures how much probability mass within the bounded continuation tree is assigned to harmful responses. However, jailbreak attacks generally do not operate on a fixed prompt. An attacker can modify, rewrite, or otherwise transform the original prompt over a sequence of queries, potentially moving the target model into regions of its input space where harmful behavior is substantially easier to elicit. Consequently, susceptibility to attack depends not only on the continuation distribution at the original prompt, but also on how readily an attacker can discover prompts with larger harmful continuation mass.

We capture this interaction through \emph{Generative-Attack Susceptibility} (\GAS{}). Let $x^{(0)}$ denote the original behavior prompt, and let
\[
    \mathcal{X}_q
    =
    \{x^{(0)},x^{(1)},\ldots,x^{(q)}\}
\]
denote the prompts considered by an attacker within a budget of $q$ queries. We define
\begin{equation}
    \GAS(q)
    =
    \max_{x\in\mathcal{X}_q}
    H_{\alpha,k,L}(x;M,J_{\phi}),
    \label{eq:gas}
\end{equation}
where $H_{\alpha,k,L}(x;M,J_{\phi})$ denotes the bounded harmful continuation mass induced by prompt $x$. Thus, \GAS{}@$q$ measures the largest harmful continuation mass exposed by the attacker within its query budget. This definition separates two distinct components of adversarial vulnerability. The \emph{outer} attack procedure determines which regions of prompt space are reached under a given search budget, while \BCA{} characterizes the target model's continuation distribution at each prompt encountered during that search. A model may therefore be robust at the original prompt yet become highly susceptible after only a small number of adversarial transformations; conversely, another model may exhibit non-negligible harmful mass at the direct prompt while changing little under additional search. Rather than assigning each model a single susceptibility score, \GAS{} represents this behavior as a function of attacker capability.

Different attack settings correspond to different constructions of $\mathcal{X}_q$. With no adversarial transformation, $\mathcal{X}_0=\{x^{(0)}\}$ and \GAS{}@0 reduces directly to \BCA{} on the original behavior prompt. For our closed library of 18 jailbreak templates, \GAS{}@18 is the maximum bounded harmful continuation mass over the transformed prompts in the library. For adaptive attackers such as \textsc{PAIR} \citep{chao2023jailbreaking} and \textsc{TAP} \citep{3737916.3739868}, \GAS{}@PAIR($q$) and \GAS{}@TAP($q$) track the running maximum over prompts discovered as the attack proceeds.

\GAS{} complements rather than replaces \ASR{}. \ASR{} records realized attack outcomes, while \GAS{} probes the underlying continuation distributions encountered during search. Reporting both as functions of query budget allows us to distinguish models that appear similar under sampled outcomes but differ substantially in the harmful behavior that remains accessible beneath those outcomes. Additionally, it provides a distributional signal for prioritizing prompts or transformations during adaptive and template-based attacks, directing additional attack queries toward regions with greater harmful continuation mass. Finally, we find that distribution-guided and outcome-guided search can expose complementary failure modes. Combining \BCA{} with outcome-based attack objectives can broaden the set of failures surfaced during red-teaming and subsequently incorporated into adversarial training.

\section{Evaluation}
\label{sec:experiments}

\subsection{Implementational Details and Experimental Setup}
\newcommand{\logoicon}[1]{%
  \raisebox{-0.28em}{\includegraphics[height=1.05em,width=1.2em,keepaspectratio]{figures/#1}}%
}

\begin{table*}[t]
\centering
\caption{
\textbf{Identical observed refusals conceal large differences in harmful reachability, and empirical ASR saturates where GAS still differentiates.}
All four models show 0\% greedy ASR (unshaded) and GAS@0 still separates them by two orders of magnitude.
Under 18 jailbreak templates, empirical ASR (fraction of behaviors where \emph{any} template ever elicits a judged-harmful completion) saturates at 100\% for three of four targets, erasing any distinction between them; $\mathrm{GAS@18}$ (the maximum bounded harmful mass over the same 18 templates) still resolves a clear gap between them. Color intensity is normalized within each shaded column; bold denotes the column maximum.
}
\label{tab:headline_results_v2}
\small
\begin{tabular}{l c c c c}
\toprule
\multirow{2}{*}{\textbf{Target model}} & \multicolumn{2}{c}{\textbf{No Attacker}} & \multicolumn{2}{c}{\textbf{Under 18-Template Attack}} \\
\cmidrule(lr){2-3} \cmidrule(lr){4-5}
& Greedy ASR & GAS@0 & Empirical ASR@18 & GAS@18 \\
\midrule
\logoicon{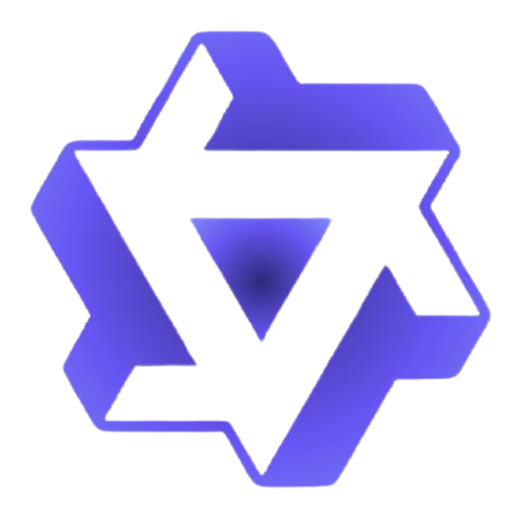}\ Qwen2.5-1.5B-Instruct
& 0\% & \cellcolor{red!1}0.004  & \cellcolor{red!31}\textbf{100.0\%} & \cellcolor{red!28}0.906 \\

\logoicon{qwen2.png}\ Qwen2.5-7B-Instruct
& 0\% & \cellcolor{red!1}0.004  & \cellcolor{red!31}\textbf{100.0\%} & \cellcolor{red!28}\textbf{0.907} \\

\logoicon{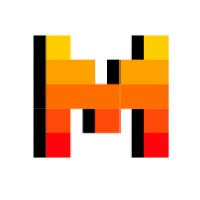}\ Mistral-7B-Instruct-v0.3
& 0\% & \cellcolor{red!28}\textbf{0.211} & \cellcolor{red!31}\textbf{100.0\%} & \cellcolor{red!24}0.782 \\

\logoicon{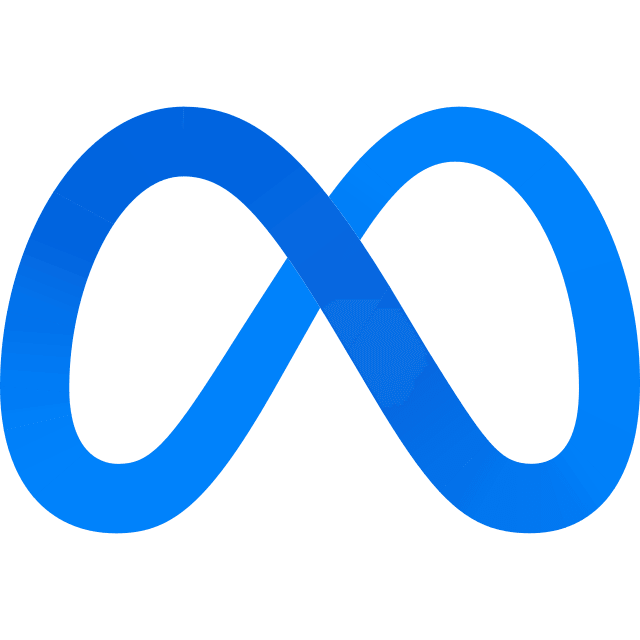}\ Llama-3.1-8B-Instruct
& 0\% & \cellcolor{red!0}0.002 & \cellcolor{red!24}76.7\% & \cellcolor{red!4}0.144 \\
\bottomrule
\end{tabular}
\end{table*}

We evaluate \BCA{} and measure \GAS{} on the 100 harmful behaviors from JailbreakBench \citep{NEURIPS2024_63092d79}. We choose JailbreakBench because it provides a standardized and widely used set of adversarial behaviors spanning diverse harm categories including harassment and discrimination, malware and cyber abuse, physical harm, fraud, disinformation, privacy violations. For our target models, we choose open weight models: \logoicon{qwen2.png} Qwen2.5-1.5B-Instruct, \logoicon{qwen2.png} Qwen2.5-7B-Instruct, \logoicon{mistralai.png} Mistral-7B-Instruct-v0.3, \logoicon{meta-color.png} Llama-3.1-8B-Instruct. These models span multiple instruction-tuned model families and sizes. These models also expose token-level probability distributions, which is essential for constructing the bounded continuation tree. 

We first evaluate \GAS{} using a fixed library of 18 jailbreak transformations spanning several common prompt-level attack strategies, including narrative and hypothetical framing, persona-based jailbreaks such as AIM, STAN, and DAN \citep{shen2024donowcharacterizingevaluating}, compliant-prefix and suffix forcing, output-format injection, and permission-based framing. We then consider two adaptive black-box attack techniques, PAIR~\citep{chao2023jailbreaking} and \textsc{TAP}~\citep{3737916.3739868}. PAIR performs sequential, response-conditioned prompt refinement, providing a simple adaptive attacker; \textsc{TAP} instead expands and prunes multiple candidate attacks at each iteration, representing a branching tree-search attacker. Together, the fixed template library, PAIR, and \textsc{TAP} allow us to study whether \BCA{} can be used to explore prompt space. We run PAIR with a single refinement stream for up to $10$ target-model queries and \textsc{TAP} with branching factor $b=2$, beam width $w=3$, and depth $d=3$, corresponding to at most $12$ target-model queries per behavior. Code and implementation details are available at \footnote{\url{https://github.com/smc879572-pixel/BCA}}.

At each of TAP's $d$ search depths, the attacker expands each of the $w$ surviving candidates into $b$ children ($b{\cdot}w$ candidates per depth) and keeps only the top-$w$ by a per-candidate score before continuing to the next depth. We call this keep/discard step \emph{pruning} throughout this section. Standard TAP scores candidates with an external judge's harm/on-topic rating. We test replacing that pruning score with each candidate's \BCA{} mass, while leaving the attacker and tree shape unchanged. This pruning score is a distinct object from the $\alpha$-$k$ bounding of \S3.2. That bounding fixes how the \emph{inner} continuation tree is built for any single candidate, and it stays identical across all three variants below. The pruning score instead decides which \emph{outer} search candidates survive to the next depth. We report on two outcome measures for the resulting trees. \emph{Any-leaf \ASR{}} measures whether at least one final leaf in the tree is considered a jailbreak, regardless of which leaf is returned. The \emph{deployment \ASR{}} measures whether the single leaf that the search's own selection rule would actually output is a jailbreak. For example, a tree with one jailbroken leaf among 12 scores as a success under any-leaf \ASR{} even if the selection rule returns one of the 11 safe leaves, a case deployment \ASR{} would score as a failure.

\subsection{Results}
\subsubsection{Do models exhibit different distributional fragility while achieving similar \ASR{}?}
\label{sec:fragility}

To answer the question, we measure (1) greedy \ASR{} and \GAS{}@0 on the 100 original JailbreakBench behaviors (i.e., how much harmful behavior is already reachable under the original unsafe prompt), and (2) \GAS{}@18 after applying a fixed library of 18 jailbreak transformations to the same behaviors (i.e., how much harmful behavior becomes reachable under adversarial reformulations of the original prompt). Table~\ref{tab:headline_results_v2} summarizes the results.  We find that greedy ASR makes all four models appear equally robust; \GAS{} sharply separates them. On direct prompts, all four models achieve greedy $0\%$ \ASR{} but their continuation distributions differ. Concretely, \GAS{}@0 is $0.211$ for Mistral-7B-Instruct-v0.3, compared to $0.004$ for both Qwen2.5-1.5B-Instruct and Qwen2.5-7B-Instruct, and $0.002$ for Llama-3.1-8B-Instruct. When the attacker is kept fixed at $q=0$, Mistral exhibits more bounded harmful mass than the other models. This is a distinction that greedy \ASR{} misses: a successful refusal does not imply that harmful continuations have been suppressed from the surrounding generation distribution. Additionally, under the 18-template attack, empirical ASR saturates at 100\% for three of the four models: Qwen2.5-1.5B, Qwen2.5-7B, and Mistral-7B-Instruct-v0.3. As a result, ASR makes these models appear equally susceptible. GAS@18, however, still distinguishes them, assigning Mistral a lower susceptibility score (0.782) than the two Qwen models (0.906 and 0.907).

\subsubsection{Does bounded harm reachability predict realized harmful behavior?} \label{sec:gas}
\begin{table}[t]
\centering
\small
\caption{\textbf{Bounded harmful reachability predicts independently judged harmful generations.}
Bounded reachability uses Qwen2.5-7B as the BCA leaf evaluator; empirical ASR is independently sampled and scored by HarmBench-Mistral-7B.
Spearman's $\rho$ and Pearson's $r$ measure rank and linear agreement between the two; Top-1 lift is the empirical-ASR gain from picking each behavior's highest-mass template over a uniform one. Darker shading denotes larger values within each metric; bold denotes the maximum}
\label{tab:validity}
\begin{tabular}{l c c c c}
\toprule
Target & $n$ pairs & Spearman $\rho$ & Pearson $r$ & Top-1 lift \\
\midrule
\logoicon{qwen2.png}\ Qwen2.5-1.5B & 540 & \cellcolor{blue!26}0.716 & \cellcolor{blue!18}0.74 & \cellcolor{blue!15}4.93$\times$ \\
\logoicon{qwen2.png}\ Qwen2.5-7B   & 540 & \cellcolor{blue!28}\textbf{0.733} & \cellcolor{blue!28}\textbf{0.83} & \cellcolor{blue!28}\textbf{7.43$\times$} \\
\logoicon{mistralai.png}\ Mistral-7B & 540 & \cellcolor{blue!11}0.611  & \cellcolor{blue!5}0.59  & \cellcolor{blue!5}2.56$\times$ \\
\logoicon{meta-color.png}\ Llama-3.1-8B & 540 & \cellcolor{blue!4}0.534 & \cellcolor{blue!23}0.79 & \cellcolor{blue!24}6.73$\times$ \\
\bottomrule
\end{tabular}
\end{table}

For each $(\text{behavior},\text{template})$ pair, we compute bounded harm reachability $\Prob[\mathbf{F}(\harm)]$ using Qwen2.5-7B-Instruct as the \BCA{} leaf evaluator. Note that we use a subset of 30 behaviors here due to the computational cost of empirical validation. We then independently sample 100 completions from the same prompt and use HarmBench-Mistral-7B classifier to estimate \ASR{}. This results in 540 $(\text{behavior},\text{template})$ pairs per model. Table~\ref{tab:validity} shows the results. We first measure the correspondence between bounded harm reachability and empirical \ASR{} using Spearman's rank correlation. We use Spearman's $\rho$ to ask whether pairs that BCA ranks as having greater harmful reachability also have higher \ASR{}. Across all models, this relationship is consistently positive: $\rho=0.716$ for Qwen2.5-1.5B, $0.733$ for Qwen2.5-7B, $0.611$ for Mistral-7B, and $0.534$ for Llama-3.1-8B. Thus, prompts assigned greater harmful probability mass by BCA tend to produce harmful completions more frequently when sampled independently. The relationship is strong enough to guide red-team allocation: choosing, for each behavior, the template ranked highest by BCA increases realized ASR by \(2.56\times\)–\(7.43\times\) relative to uniform template selection.

Because each behavior is evaluated under multiple templates and each template is applied across multiple behaviors, we complement Spearman's correlation with a mixed-effects logistic regression that accounts for both sources of variation~\citep{gelman2007data}. We found that some behaviors are more susceptible to harmful responses, and some templates are more successful at eliciting harmful completions. Nevertheless, bounded harm reachability remains positively associated with harmful generation after accounting for these differences.

\subsection{adaptive red-teaming with BCA}
\label{sec:tap}
\begin{table*}[t]
\centering
\caption{
\textbf{BCA is useful for search across model families, but outcome-aligned scoring remains necessary for selection.}
\textbf{(a)} BCA-based pruning matches or exceeds judge-based pruning on most targets, but selecting with BCA alone collapses deploy ASR on every target (worst for DeepSeek-R1, $-$41.6 points); selecting with the outcome judge instead recovers full deploy performance.
\textbf{(b)} Judge- and BCA-guided searches find partially disjoint jailbreaks; running both raises coverage on all four models, most on Qwen2.5-7B (self-attack), whose 48.3\% neither-succeeds rate is the table's highest. Shading is scaled independently per panel: \textbf{(a)} the \hlred{any-leaf-to-deploy collapse} when BCA also selects the deployed candidate; \textbf{(b)} the \hlblue{share of paired runs where neither search succeeds}.
}
\label{tab:search_selection_complementarity_four}
\footnotesize
\setlength{\tabcolsep}{4pt}

\textbf{(a) Search signal versus deployment-selection signal}\\[2pt]
\begin{tabular}{l cc cc cc cc}
\toprule
& \multicolumn{2}{c}{\logoicon{meta-color.png} Llama-3.1-8B} & \multicolumn{2}{c}{\logoicon{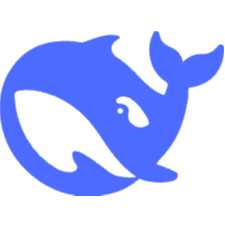} DeepSeek-R1} & \multicolumn{2}{c}{\logoicon{qwen2.png} Qwen2.5-7B$^\dagger$} & \multicolumn{2}{c}{\logoicon{mistralai.png} Mistral-7B} \\
\cmidrule(lr){2-3} \cmidrule(lr){4-5} \cmidrule(lr){6-7} \cmidrule(lr){8-9}
Search $\times$ Select & Any-leaf & Deploy & Any-leaf & Deploy & Any-leaf & Deploy & Any-leaf & Deploy \\
\midrule
Judge $\times$ Judge & 57.0\% & 57.0\% & 64.2\% & 64.2\% & 33.3\% & 33.3\% & 65.8\% & \textbf{65.8}\% \\
BCA $\times$ BCA      & 58.3\% & \cellcolor{red!23}24.1\% & 73.3\% & \cellcolor{red!31}31.7\% & 38.3\% & \cellcolor{red!17}17.5\% & 62.5\% & \cellcolor{red!36}36.7\% \\
BCA $\times$ Judge     & 58.3\% & \textbf{58.3\%} & 73.3\% & \textbf{73.3\%} & 38.3\% & \textbf{38.3\%} & 62.5\% & 62.5\% \\
\bottomrule
\end{tabular}

{\scriptsize $^\dagger$ self-attack.}

\vspace{8pt}
\textbf{(b) Complementary deployable jailbreak coverage}\\[2pt]
\begin{tabular}{l cccc}
\toprule
Outcome & \logoicon{meta-color.png} Llama & \logoicon{deepseek.png} DeepSeek & \logoicon{qwen2.png} Qwen$^\dagger$ & \logoicon{mistralai.png} Mistral \\
& ($N{=}120$) & ($N{=}120$) & ($N{=}120$) & ($N{=}120$) \\
\midrule
Both succeed       & 37.3\% & 55.8\% & 20.0\% & 54.2\% \\
Judge-guided only  & 19.6\% & 8.3\%  & 13.3\% & 11.7\% \\
BCA-guided only    & 21.6\% & 17.5\% & 18.3\% & 8.3\%  \\
Neither succeeds   & \cellcolor{blue!11}21.6\% & \cellcolor{blue!9}18.3\% & \cellcolor{blue!24}48.3\% & \cellcolor{blue!13}25.8\% \\
\midrule
\rowcolor{gray!12}
\textbf{Union: either succeeds} & \textbf{78.4\%} & \textbf{81.7\%} & \textbf{51.7\%} & \textbf{74.2\%} \\
\bottomrule
\end{tabular}
\end{table*}

We also evaluate the utility of \BCA{}{} as an objective for discovering attacks by adaptive red-teaming methods. Specifically, we integrate \BCA{} into \textsc{PAIR} and \textsc{TAP} by replacing the judge-based candidate-pruning score with \BCA{}. We evaluate all 7B-scale models defined in Section 4.1 and additionally include DeepSeek-R1-Distill-Llama-8B.  Unless otherwise noted, we use the exact bounded continuation process as above ($\alpha=0.99$, $k=2$, $L=8$) and the Qwen2.5-7B semantic judge. Then, we compare \emph{judge-guided} \textsc{TAP} against \emph{\BCA{}-guided} \textsc{TAP}. We define two metrics for measuring attack success. \emph{Any-leaf \ASR{}} measures whether the search tree contains at least one successful jailbreak: a run counts as successful if \emph{any} explored leaf receives a 10/10 score from the outcome judge, irrespective of which leaf the search procedure ultimately selects. In contrast, \emph{deploy \ASR{}} evaluates only the single leaf returned by the attack and measures whether the candidate actually selected for deployment is a successful jailbreak. For the sake of brevity, the results for integrating \BCA{} with PAIR are provided in Appendix \ref{app:comp}.

\textbf{BCA-guided and judge-guided search expose complementary failures.} Across 120 runs each for four target models, BCA-guided TAP discovers successful jailbreaks at rates comparable to judge-guided TAP, and exceeds it on three of the four targets in any-leaf ASR (Table \ref{tab:search_selection_complementarity_four}a).  More importantly, the two search signals uncover partially disjoint sets of failures (Table \ref{tab:search_selection_complementarity_four}b). Across all four targets, some jailbreaks are found only by judge-guided search and others only by BCA-guided search. Taking the union raises deployable jailbreak coverage to 78.4\% on Llama-3.1-8B, 81.7\% on DeepSeek-R1-Distill-Llama-8B, 51.7\% on Qwen2.5-7B, and 74.2\% on Mistral-7B. This complementarity suggests that BCA captures information about the target model's continuation distribution that is not redundant with an outcome-based judge, making the two signals useful in combination when the goal is to broaden failure coverage rather than optimize a single attack trajectory.

\section{Related work}
\label{sec:related}

Early hand-written, one-shot jailbreaks were standardized by HarmBench~\citep{harmbench} and JailbreakBench~\citep{NEURIPS2024_63092d79} into fixed behavior sets with automated pass/fail judges. Attacks evolved into adaptive search using the target's own responses as feedback---PAIR \citep{chao2023jailbreaking} and \textsc{TAP} \citep{3737916.3739868}, part of the broader automated red-teaming paradigm \citep{perez2022redteaminglanguagemodels}, including genetic and fuzzing-based search over wider candidate sets \citep{liu2024autodan, 3698900.3699161}---but evaluation remained a binary label, aggregated across a query budget. While later work refines what this label should reward \citep{strongreject}.\footnote{Appendix~\ref{app:scoring} surveys additional work refining single-response jailbreak scoring.} These methods differ in how they score a sampled response, using binary labels, severity ratings, or weighted sub-question decompositions, but the object being evaluated remains a single observed completion. We show this framing cannot separate two models with identical greedy outputs, but whose nearby harmful mass differs by two orders of magnitude (\S\ref{sec:experiments}). 

A second limitation of prior work is that ASR numbers are also not comparable across papers: \citet{chouldechova2026comparisonrequiresvalidmeasurement} show ASR shifts once threat models, judges, or budgets differ. Closest to our work is \citet{maple2026} who report a distribution over an attack's \emph{hyperparameter configurations}---in contrast, we report one over the \emph{target model's own stochastic continuations} under one fixed prompt, well defined under explicit bounding and evaluator assumptions.

Beyond measurement, \citet{jailbroken} attribute this gap to safety training failing via two mechanisms: competing objectives between helpfulness and harmlessness, and mismatched generalization between training and capability. \citet{ICLR2025_88be0230} sharpen this by localizing the vulnerability to the first few output tokens: safety training reshapes that prefix exclusively, so a model can refuse on the surface while retaining high harmful mass just past it. Prefilling \citep{li2025prefillleveljailbreakblackboxrisk} and logprob-targeted attacks \citep{ICLR2025_63fa7efd} exploit this directly, manipulating initial token probabilities to bypass search entirely. \BCA{} operationalizes this phenomenon as a diagnostic, not an attack---our refusal-collapse finding turns a qualitative account of shallow alignment into a measured probability.

Closest to our approach is LLMChecker \citep{gross2025boundedpctlmodelchecking}, which introduced \(\alpha\)-\(k\)-bounded PCTL model checking for LLM generation. We build directly on this formulation but extend it along three dimensions. First, we replace recursive per-node tree expansion and external Storm-based model checking with depth-wise GPU-batched expansion and in-process sparse backward induction, reducing the number of sequential model-inference rounds from \(O(k^L)\) to \(O(L)\) while retaining the same worst-case \(O(k^L)\) bounded tree size. Second, we extend the checked properties from primarily deterministic lexical and structural attributes to behavior-conditioned semantic harm evaluated by pretrained models. Third, we compose bounded reachability with adaptive red teaming: \GAS{} tracks harmful reachability over an attacker's query budget, and \BCA{}serves as a search signal within PAIR and TAP.

\section{Conclusion}

Current jailbreak evaluations largely reduce model safety to an observed outcome: whether an attack produces a harmful response. In this work, we argue that this view is incomplete. Autoregressive models define distributions over possible continuations, and a safe decoded response can coexist with substantial harmful probability mass elsewhere in the model’s continuation distribution. We introduced \BCA{}, a bounded measure of harmful reachability within this continuation distribution, and \GAS{}, which extends this perspective to adaptive attacks by tracking how harmful reachability evolves with attacker budget. Across models and JailbreakBench behaviors, these quantities reveal safety differences that are invisible to greedy \ASR{}, correlate with independently measured attack success, and expose failures that are complementary to those found by outcome-guided search. In particular, distribution-guided and outcome-guided attacks often uncover different jailbreaks, making the continuation distribution useful not only for diagnosis but also for broadening red-team coverage. By making latent harmful reachability measurable and operational, \BCA{} provides a step toward safety evaluations that reason about the generation process itself rather than only its sampled outcomes.

\section*{AI use statement}

In this work, we \textbf{have used} generative AI tools to design and provide feedback on experiments and implement methods.
We have \textbf{not used} generative AI tools to help develop theoretical models or conceptual frameworks, help in the writing of proofs, propose or refine hypotheses, assist with translation, support qualitative and thematic data analysis or the interpretation of results. 
The generation of synthetic data sets, the formulation of mathematical claims, the provision of critical ingredients for proving mathematical claims, and the cleaning and reformatting of the dataset are \textbf{not applicable} to this work.

Additionally, we used generative AI tools to create Figure \ref{fig:tap-overlap}, suggest experimental parameters, create and edit software code, create artifacts, draft parts of the research paper, analyze existing literature, source/search for information, and edit the paper to improve readability.

We have reviewed all AI-assisted work: figures and artifacts were checked against the experimental data before inclusion, AI-suggested experimental parameters were reviewed against pilot results, AI-assisted code was tested and verified for correctness by two of the authors, and AI-drafted or AI-edited text was reviewed by all authors for factual accuracy. We take responsibility for the final content of this work, including text, claims or artifacts produced with the aid of generative AI.

\section*{Ethics statement} 
\label{sec:ethics}
This work studies adversarial prompts and harmful behaviors to improve safety evaluation. We report template \emph{families} rather than operational jailbreak strings, follow public benchmark protocols, and emphasize defensive use: triage, auditing, and adversarial training coverage. Dual-use risk remains; we recommend gated release of attack traces consistent with JailbreakBench norms.

\section*{Reproducibility Statement}
All targets and judges used in this work are open-weight models (Qwen2.5-1.5B/7B-Instruct, Mistral-7B-Instruct-v0.2/v0.3, Llama-3.1-8B-Instruct, HarmBench-Mistral-7B, Llama Guard~3), and all behaviors are drawn from the public JailbreakBench suite, so every experiment in this paper can be rerun without access to closed APIs. Algorithm~\ref{alg:bca} gives the exact bounded-reachability computation, and \S\ref{sec:experiments} reports the $\alpha$, $k$ and $L$ used in each experiment, along with attacker budgets for the closed-template library, PAIR, and \textsc{TAP}. Consistent with JailbreakBench release norms and our ethics statement (\S\ref{sec:ethics}), we release our code for constructing and evaluating bounded continuation trees, along with behavior categories and aggregate results, at an anonymized repository: \url{https://github.com/smc879572-pixel/BCA.git}. A de-anonymized version will be made available upon acceptance. We withhold full jailbreak prompts, template text, and attack traces to limit dual-use risk. We hope this transparency will allow the community to verify our findings and build upon our work, for instance by auditing additional models' hidden harmful mass or extending \BCA{} to new judges and attackers. Random seeds are reported wherever they affect results (seeds 42--45 for the multi-seed \textsc{TAP} experiments).

Two considerations limit the absolute reproducibility of our experiments. First, all reported reachability probabilities are conditional on the specific judge checkpoints used (Qwen2.5-7B-Instruct, HarmBench-Mistral-7B, Llama Guard 3). Reproducing our exact numbers requires picking the same model versions, and results may shift if providers update or deprecate these checkpoints. Second, our experiments involve probabilistic target models which introduce variance into our results.

\bibliography{iclr2027_conference}
\bibliographystyle{iclr2027_conference}

\appendix

\section{Related Work on Single-response Jailbreak Scoring}
\label{app:scoring}
Section \ref{sec:related} reviews the most directly related prior work, focused on jailbreak evaluation and ASR. We now cover other related prior work that has refined single-response jailbreak scoring.

Beyond StrongREJECT \citep{strongreject}, several works refine what counts as success on a single sampled response, without moving to a distributional view. AttackEval~\citep{attackeval} proposes a coarse- and fine-grained scoring framework in the same view as sub-question decomposition methods. \citet{rethinking_eval} argue that binary success/fail conflates three distinct axes: whether a safeguard was violated, whether the response was substantive enough to be useful, and whether harmful content was accurate rather than merely harmful-sounding. They propose individual scoring along each axis. \citet{amess} show that ASR-based rankings are only weakly aligned with a defender's utility from an attack, a distinct critique of ASR: ASR misleads \emph{prioritization} decisions, not just severity judgments.

\section{Harmful Reachability via Backward Induction}
\label{app:induction}
After constructing the bounded continuation tree to depth $L$, an evaluator $J$ assigns each leaf state $s$ a binary harm label,
\begin{equation}
J(s)\in\{0,1\},
\end{equation}
where $J(s)=1$ denotes a harmful continuation. We define a value function $V:S\rightarrow[0,1]$ over the bounded tree. For each leaf state,
\begin{equation}
V(s)=J(s).
\end{equation}
For each internal state, we compute its value by backward induction:
\begin{equation}
V(s)
=
\sum_{y\in\mathcal{T}_{\alpha,k}(s)}
\widetilde{P}(s,s\circ y)\,
V(s\circ y),
\label{eq:backward_induction}
\end{equation}
where $\widetilde{P}$ denotes the transition probabilities of the bounded DTMC. Evaluating Eq.~\ref{eq:backward_induction} from depth $L$ back to the root gives
\begin{equation}
H_{\alpha,k,L}(x;M,J)
=
V(s_0),
\qquad s_0=x,
\label{eq:bca_reachability}
\end{equation}
the exact probability of reaching a harmful leaf under the bounded DTMC.

Because the bounded continuation structure is a tree, these values can be computed in a single reverse traversal. Once the tree has been constructed and labeled, backward induction therefore requires $O(|S|)$ time, where $|S|$ is the number of states in the bounded tree.

\begin{algorithm}[ht]
\caption{Exact \BCA{} on an $\alpha$-$k$ bounded continuation tree}
\label{alg:bca}
\begin{algorithmic}[1]
\Require prompt $x$, model $M$, evaluator $J$, horizon $L$, bounds $\alpha,k$
\State Construct the $\alpha$-$k$ bounded tree rooted at $s_0=x$ to depth at most $L$
\State Label each leaf $s$ with $\ell(s)\gets J(s)$
\State $V(s)\gets \mathbf{1}[\ell(s)=\harm]$ for every leaf $s$
\For{$d=L-1$ \textbf{downto} $0$}
    \ForAll{nonterminal states $s$ at depth $d$}
        \State $V(s)\gets
        \displaystyle\sum_{t\in T_{\alpha,k}(s)}
        \widetilde P(s,s\!\cdot\! t)\,V(s\!\cdot\! t)$
    \EndFor
\EndFor
\State \Return $V(s_0)$
\end{algorithmic}
\end{algorithm}

\section{Implementation}
\label{app:impl}

Our GPU implementation uses breadth-first batched inference (vLLM/HuggingFace backends) and sparse backward induction for exact trees.
Code and interactive complementarity visualizations are available at an anonymized repository: \url{https://github.com/smc879572-pixel/BCA.git}.

\section{BCA-GUIDED SEARCH WITH PAIR}
\label{app:comp}
We provide additional results evaluating BCA as a search signal within PAIR. The same qualitative findings observed with TAP persist under sequential prompt refinement. BCA-guided PAIR discovers successful jailbreaks at rates comparable to judge-guided PAIR on three of the four target models, while underperforming on Mistral-7B. The two search signals also expose partially disjoint failures on every target. Taking their union increases deployable jailbreak coverage to 66.0\% on Llama-3.1-8B, 67.1\% on DeepSeek-R1-Distill-Llama-8B, 31.6\% on Qwen2.5-7B, and 65.0\% on Mistral-7B, indicating that the complementarity between distribution-guided and outcome-guided search is not specific to TAP.

\begin{table*}[!t]
\centering
\caption{
\textbf{The search-vs-selection and complementary-coverage findings replicate under PAIR, not just \textsc{TAP}.}
\textbf{(a)} BCA-guided search discovers jailbreaks at a rate comparable to judge-guided search on three of four targets; Mistral-7B is the one target where it underperforms outright, consistent with the same direction seen under \textsc{TAP}. Selecting with BCA alone discards most successful candidates on every target; selecting with the outcome judge instead recovers deploy performance. Shading marks the \hlred{any-iteration-to-deploy drop}, scaled across all four models; \textbf{bold} marks the recovered value.
\textbf{(b)} Judge- and BCA-guided PAIR chains find partially disjoint failures on every target; running both raises coverage over either alone, even on Qwen2.5-7B, the most PAIR-resistant of the four. Shading marks the \hlred{Neither-succeeds rate}, scaled across all four models.
}
\label{tab:pair_search_selection_complementarity}
\footnotesize
\setlength{\tabcolsep}{4pt}

\textbf{(a) Search signal versus deployment-selection signal (PAIR, $q{=}10$)}\\[2pt]
\begin{tabular}{l cc cc cc cc}
\toprule
& \multicolumn{2}{c}{\logoicon{meta-color.png} Llama-3.1-8B$^\dagger$} & \multicolumn{2}{c}{\logoicon{deepseek.png} DeepSeek-R1$^\ddagger$} & \multicolumn{2}{c}{\logoicon{qwen2.png} Qwen2.5-7B$^\dagger$} & \multicolumn{2}{c}{\logoicon{mistralai.png} Mistral-7B$^\ddagger$} \\
\cmidrule(lr){2-3} \cmidrule(lr){4-5} \cmidrule(lr){6-7} \cmidrule(lr){8-9}
Search $\times$ Select & Any-iter. & Deploy & Any-iter. & Deploy & Any-iter. & Deploy & Any-iter. & Deploy \\
\midrule
Judge $\times$ Judge & 56.9\% & 56.9\% & 48.1\% & 48.1\% & 20.3\% & 20.3\% & 53.8\% & 53.8\% \\
BCA $\times$ BCA      & 51.7\% & \cellcolor{red!28}30.0\% & 46.2\% & \cellcolor{red!24}27.5\% & 21.2\% & \cellcolor{red!10}13.8\% & 43.8\% & \cellcolor{red!21}27.5\% \\
BCA $\times$ Judge     & 51.7\% & \textbf{51.7\%} & 46.2\% & \textbf{46.2\%} & 21.2\% & \textbf{21.2\%} & 43.8\% & \textbf{43.8\%} \\
\bottomrule
\end{tabular}

{\scriptsize $^\dagger$ self-attack. \quad $^\ddagger$ external Qwen2.5-7B attacker.}

\vspace{8pt}
\textbf{(b) Complementary deployable jailbreak coverage}\\[2pt]
\begin{tabular}{l cccc}
\toprule
Outcome & \logoicon{meta-color.png} Llama & \logoicon{deepseek.png} DeepSeek & \logoicon{qwen2.png} Qwen  & \logoicon{mistralai.png} Mistral \\
\midrule
Both searches succeed & 38.0\% & 26.6\% & 10.1\% & 32.5\% \\
Judge-guided only      & 16.0\% & 21.5\% & 10.1\% & 21.2\% \\
BCA-guided only         & 12.0\% & 19.0\% & 11.4\% & 11.2\% \\
Neither succeeds        & \cellcolor{red!14}34.0\% & \cellcolor{red!13}32.9\% & \cellcolor{red!28}68.4\% & \cellcolor{red!14}35.0\% \\
\midrule
\rowcolor{gray!12}
\textbf{Union: either succeeds} & \textbf{66.0\%} & \textbf{67.1\%} & \textbf{31.6\%} & \textbf{65.0\%} \\
\bottomrule
\end{tabular}
\end{table*}

\end{document}